# Estimation of River Water Surface Elevation Using UAV Photogrammetry and Machine Learning

Radosław Szostak[a,*], Marcin Pietroń[b], Przemysław Wachniew[a], Mirosław Zimnoch[a], Paweł Ćwiąkała[c]

[a] *AGH University of Science and Technology, Faculty of Physics and Applied Computer Science, ul. Reymonta 19, building D-10, Kraków, Poland*
[b] *AGH University of Science and Technology, Faculty of Computer Science, Electronics and Telecommunications, al. Mickiewicza 30, building D-17, Krakow, Poland*
[c] *AGH University of Science and Technology, Faculty of Geo-Data Science, Geodesy, and Environmental Engineering, al. Mickiewicza 30, building C-4, Krakow, Poland*

**Abstract**

Unmanned aerial vehicle (UAV) photogrammetry allows for the creation of orthophotos and digital surface models (DSMs) of a terrain. However, DSMs of water bodies mapped with this technique reveal water surface distortions, preventing the use of photogrammetric data for accurate determination of water surface elevation (WSE).

Firstly, we propose a new solution in which a deep learning (DL) convolutional neural network (CNN) is used as a WSE estimator from photogrammetric DSMs and orthophotos. Second, we improved the previously known "water-edge" method by filtering the outliers using a forward-backwards exponential weighted moving average (FBEWMA). An additional improvement in these two methods was achieved by performing a linear regression of the WSE values against chainage. Furthermore, the solutions estimate the uncertainty of the predictions, which allows the evaluation of the model credibility when the ground-truth values are unknown.

To our knowledge, this is the first approach in which DL was used for this task. A brand new machine learning data set has been created for the purpose of this work. It was collected on a small lowland river in winter and summer conditions. It consists of 322 samples, each corresponding to a 10 by 10 meter area of the river channel and adjacent land. Each data set sample contains orthophoto and DSM arrays as input, along with a single ground-truth WSE value as output. The data set was supplemented with data collected by other researchers that compared the state-of-the-art methods for determining WSE using an UAV. This makes these additional data an excellent benchmark to compare the method developed with existing ones.

The results of the DL solution were verified using the reliable k-fold cross-validation method, in which samples from a single survey campaign were excluded from training and used as a validation subset. This provided an in-depth examination of the model's ability to perform on previously unseen data. The WSE RMSE values differ for each k-fold cross-validation subset and range from 1.7 cm up to 17.2 cm. The results obtained are remarkably good considering the small size of the training data set.

The RMSE results of the improved "water-edge" method are ranging from 0.4 cm to 10.1 cm, and on average they are at least six times lower than the RMSE results achieved by the conventional "water-edge" method.

The results obtained by new DL-based and improved "water-edge" methods are predominantly outperforming existing methods using photogrammetric products. Moreover, the result obtained by the DL based method proved to be better than any other UAV-based measurement method (RADAR, LIDAR, photogrammetry) tested in the same case study.

---

\* Corresponding author

# 1. Introduction

The management of water resources constitutes one of central issues of the sustainable development for the environment and human health. The sustainable use of water resources relies on understanding the complex and interrelated processes that affect the quantity and quality of water available for human needs, economic activities, and ecosystems. Global demand for freshwater continues to increase at a rate of 1% per year since 1980 driven by population growth and socioeconomic changes. Simultaneously, the increase in evaporation caused by rising temperatures leads to a decrease in streamflow volumes in many areas of the world, which already suffer from water scarcity problems. Achieving socioeconomic and environmental sustainability under such challenging conditions will require the application of innovative technologies, capable of measuring hydrological characteristics at a range of spatial and temporal scales (Blöschl et al., 2019). Traditional surface water management practices are primarily based on data collected from networks of in situ hydrometric gauges. Point measurements do not provide sufficient spatial resolution to fully characterize river networks. Moreover, the decline of existing measurement networks is being observed all over the world and many developing regions lack them altogether (Lawford et al., 2013). Remote sensing methods are considered a solution to cover data gaps specific to point measurement networks (McCabe et al., 2017). A leading example of remote sensing are measurements made from satellites. However, due to the too low spatial resolution, satellite data are suitable only for studying the largest rivers. For example, the SWOT mission will allow only observation of rivers of width greater than 50-100 m (Pavelsky et al., 2014). Small surface streams of the first and second order (according to Strahler's classification (Strahler, 1957)) constitute 70%-80% of the length of all rivers in the world. Small streams play a significant role in hydrological systems and provide an ecosystem for living organisms (Wohl, 2017). In this regard, measurement techniques based on unmanned aerial vehicles (UAVs) are promising in many key aspects, as they provide observations in high spatial and temporal resolution, their deployment is simple and fast, and can be used in inaccessible locations (Vélez-Nicolás et al., 2021). One of the most important river characteristics is spatially distributed water surface elevation (WSE), as it is used for the validation and calibration of hydrologic, hydraulic, or hydrodynamic models to make hydrological forecasts, including predicting dangerous events such as floods and droughts (Asadzadeh Jarihani et al., 2013; Domeneghetti, 2016; Langhammer et al., 2017; Montesarchio et al., 2014; Tarpanelli et al., 2013).

Photogrammetric Structure-from-Motion (SfM) algorithms are able to generate orthophotos and digital surface models (DSMs) of terrain from multiple aerial photographs. Photogrammetric DSMs are precise in determining the elevation of solid surfaces to within a few centimeters (Bühler et al., 2017; Ouédraogo et al., 2014), but water surfaces are usually falsely stated. This is related to the fact that the general principle of SfM algorithms is based on the automatic search for distinguishable and static terrain points that appear in several images showing these points from different perspectives. The surface of water lacks such points as it is uniform, transparent, and in motion. The transparency of the water makes the surface level of the river depicted on the photogrammetric DSM lower than in reality. The river bottom is represented by photogrammetric DSMs for clear and shallow streams (Kasvi et al., 2019). Photogrammetric DSMs for opaque water bodies are affected by artifacts brought on by lack of distinguishable key points (Woodget et al., 2014). (Woodget et al., 2014), (Javernick et al., 2014) and (Pai et al., 2017) demonstrated that it is possible to read the WSE from photogrammetric DSM at shorelines ("water-edge") where the river is shallow, so there are no undesirable effects associated with light penetration below the water surface. However, (Bandini et al., 2020) proved that this method gives satisfactory results only for unvegetated and smoothly sloping shorelines where the boundary line between water and land is easy to define. For this reason, this method is not suitable for universal automation.

The exponentially growing interest (Pugliese et al., 2021) and the impressive results of machine learning algorithms in various fields offer promising prospects for the development and application of this technology in hydrological sciences. For this reason, in this article, we propose a new method based on deep neural networks that allows the estimation of the WSE of small rivers from photogrammetric products. We also present an improved water-edge method and an uncertainty estimation based on exponential moving average. The results of the methods presented are compared with the results achieved in previous studies on remote sensing of WSE measurements in small rivers.

## 2. Materials and methods

*2.1. Data*

*2.1.1. Case study site*

Photogrammetric data and WSE observations were obtained for Kocinka, a small lowland river (length 40 km, catchment area 260 km$^2$) located in the Odra River basin in southern Poland. Data were collected on two river stretches with similar hydromorphological characteristics and different water transparency:

i. ca. 700 m stretch of the Kocinka river located near Grodzisko village (50.8744N, 18.9711E). This stretch has a WS width of about 2 m. There are no trees in close proximity to the river. The riverbed is made up of dark silt and the water is opaque. The banks and the riverbed are overgrown with rushes that protrude above the water surface. The banks are steeply sloping at angles of ca. 50 ° to 90 ° relative to the water surface. There are marshes nearby, with river water flowing into them in places. Data from this stretch were collected on the following days:

   a. December 19, 2020 – Total cloud cover was present during the measurements. Due to the winter season, the foliage was reduced. Samples obtained from this survey are labeled with the identifier "GRO20".

   b. July 13, 2021 – There was no cloud cover during the measurements. The rushes were high and the water surface was densely covered with Lemna plants. Samples obtained from this survey are labeled with the identifier "GRO21".

ii. ca. 700 m stretch of the Kocinka river located near Rybna village (50.9376N, 19.1143E). This stretch has a WS width of about 3m and is overhung by sparse deciduous trees. There is a pale, sandy riverbed that is visible through the clear water. There are no rushes that emerge from the riverbed. The banks slope at angles of about. 20 ° to 90 ° relative to the water surface. Data from this stretch were collected on the following days:

   a. December 19, 2020 – Total cloud cover was present during the measurements. Due to the winter season, the trees were devoid of leaves and the grasses were reduced. Samples obtained from this survey are labeled with the identifier "RYB20".

   b. July 13, 2021 – There was no cloud cover during the measurements. The offshore grasses were high. With good lighting and exceptionally clear water, the riverbed was clearly visible through the water. The samples obtained from this survey are labeled with the identifier "RYB21".

The orthophotos of the Grodzisko and Rybna case studies are shown in Figure 1. The photo of part of the Rybna case study is shown in Figure 2.

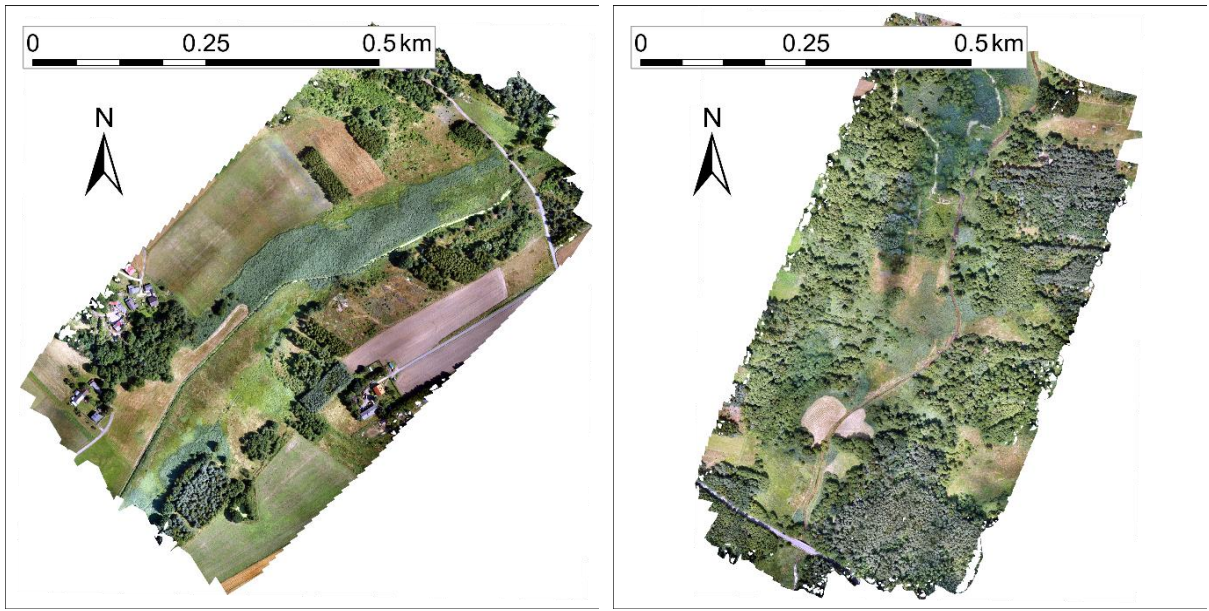

Figure 1. Orthophoto imagery from Grodzisko (left) and Rybna (right) case studies from 2021.

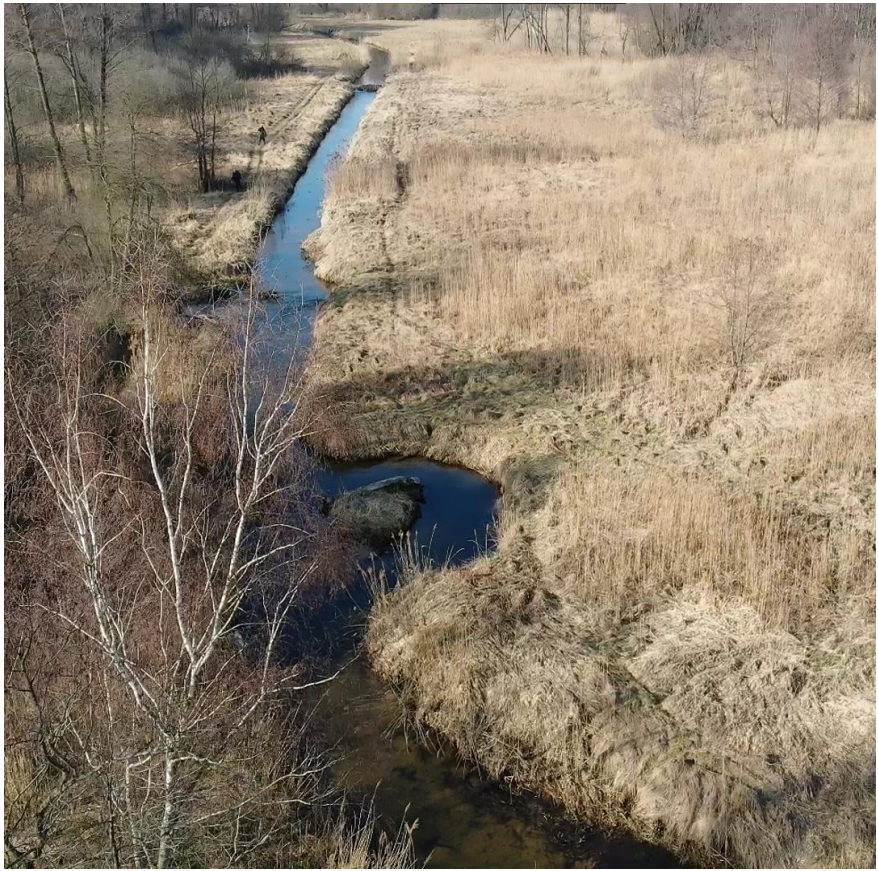

Figure 2. Part of the Kocinka River in Rybna stretch (March 2022).

Furthermore, the data set was supplemented with data from surveys conducted by (Bandini et al., 2019) over approximately 2.3km stretch of the river Åmose Å (Denmark) on November 21, 2018. The river is channelized and well maintained. The banks are overgrown with low grass and the neighboring few trees are devoid of leaves due to winter. (Bandini et al., 2020) describe further details about this case study in their publication. This research tested current state-of-the-art methods to measure river WSE with UAVs using radar, lidar and photogrammetry-based measurements. This makes the data set acquired there an excellent benchmark for evaluating the accuracy of the new method by comparing it with existing methods. The samples obtained from this survey are labeled in our data set with the identifier "AMO18".

*2.1.2. Field surveys*

During the survey campaigns, photogrammetric measurements were conducted over the river area. Aerial photos were taken from a DJI S900 UAV using a Sony ILCE a6000 camera with a Voigtlander SUPER WIDE HELIAR VM 15 mm f/4.5 lens. The flight altitude was approximately 77 m AGL, resulting in a 20 mm terrain pixel. The front overlap was 80%, and side overlap was 60%. In addition to drone flights, Ground Cotrol Points (GCPs) were established homogeneously in the area of interest using a Leica GS 16 RTN GNSS receiver. Ground-truth WSE point measurements were also made using an RTN GNSS receiver. They were carried out along the river every approximately 10-20 meters on both banks.

*2.1.3. Data processing*

Orthophoto and DSM raster files were generated using Agisoft Metashape photogrammetric software. GCPs were used to embed rasters in a geographic reference system of latitude, longitude, and elevation. Further data processing was performed using ArcGIS ArcMap software. Each of the obtained rasters had the width and height of several tens of thousands of pixels and represented a part of a basin area exceeding 30ha. For the machine learning application, samples representing 10m x 10m areas of the terrain were manually extracted from large-scale Orthophoto and DSM rasters. Each sample contains areas of water and adjacent land. The samples do not overlap.

The point measurements of ground-truth WSE were interpolated using polynomial regression as a function of chainage along the river centerline. In the situation that the beaver dam caused an abrupt change in the WSE, regressions were made separately for the sections upstream and downstream of a dam. The WSE values interpolated by regression analysis were assigned to the raster samples according to the geospatial location. The average WSE values from a river centerline segment located within the sample area were assigned to the sample as ground-truth WSE. The Standard Error of Estimate metric (Siegel, 2016) was used to determine the accuracy of ground-truth data. It was calculated using the formula:

$$S_e = \sqrt{\frac{1}{n-2} \sum_{i=1}^{n} (WSE_i - \widehat{WSE}_\iota)^2}, \qquad (1)$$

where:
$n$ – number of WSE point measurements,
$WSE_i$ – measured WSE value,
$\widehat{WSE}_\iota$ – WSE from regression analysis

The results of the Standard Error of Estimate examination are included in Table 1. We see that the ground-truth WSE error is up to 2 cm.

Table 1. Statistics of the subsets of machine learning data sets acquired for each survey subset.

| Subset ID | Number of WSE point measurements | Standard Error of Estimate for ground-truth regression (m) | Number of extracted data set samples |
|---|---|---|---|
| GRO21 | 36 | 0.012 | 64 |
| RYB21 | 52 | 0.013 | 55 |
| GRO20 | 84 | 0.020 | 72 |
| RYB20 | 76 | 0.016 | 57 |
| AMO18 | 7235 | 0.020 | 74 |

Figure 3 shows a diagram depicting the steps in data set preparation that include both fieldwork and data processing.

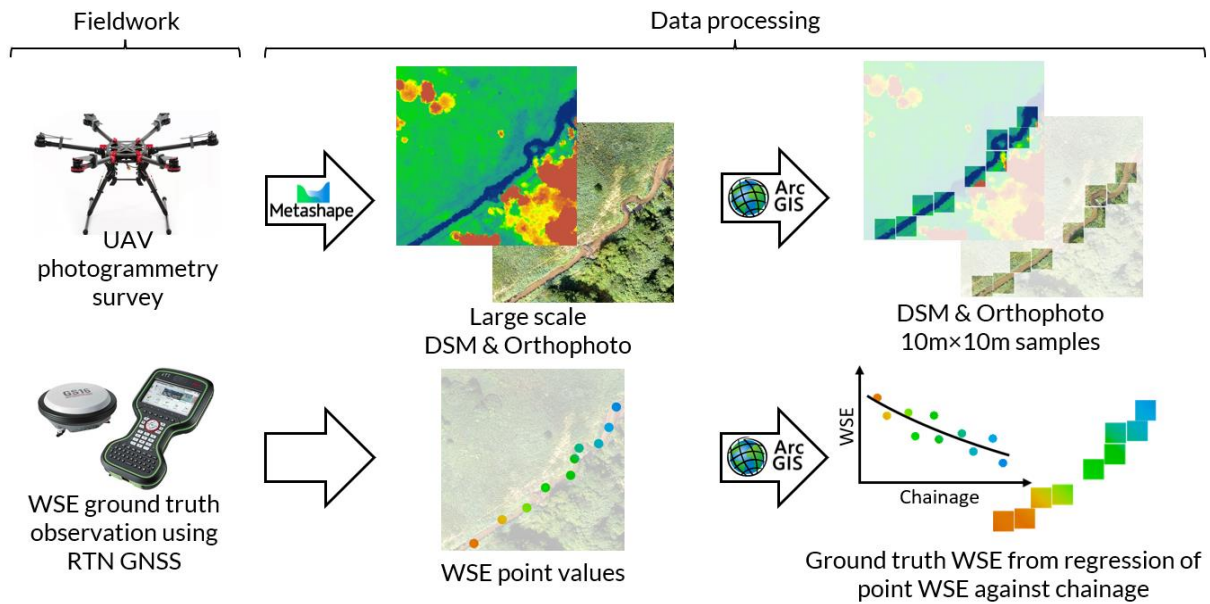

Figure 3. Schematic representation of the workflow for preparing the data set.

### 2.1.4. Machine learning data set structure

The machine learning data set comprises 322 samples. For details on the number of samples in each subset, see Table 1. Every sample includes the data described below.

- **Photogrammetric orthophoto** - a square crop of an orthophoto representing 10m × 10m area, containing the water body of a river and adjacent land. Grayscale image represented as a 256 × 256 array of integer values from 0 to 255 (1-channel image of 256 × 256 pixels).

- **Photogrammetric DSM** – a square crop of the DSM representing the same area as the Orthophoto sample described above. Stored as 256 × 256 array of floating point numbers containing elevations of pixels expressed in m MSL.

- **Water Surface Elevation** – ground-truth WSE of the water body segment included in Orthophoto and DSM samples. Represented as a single floating point value expressed in m MSL.

- **Metadata** – the following additional information is stored for each sample:

- **Mean**, standard deviation, minimum, and maximum values of the photogrammetric DSM sample array, which can be used for standardization or normalization. Represented as floating point values expressed in m MSL.
- **Centroid latitude and longitude** – WGS-84 geographical coordinates of the centroid of the shape of the sample area. Represented as floating-point numbers.
- **Chainage** – the centroid of the position of the sample according to the relative chainage of a given river section. Used for linear regression, increasing accuracy, and providing uncertainty. For additional information, see section 2.2.6).
- **Subset ID** – text value that identifies the survey subset to which the sample belongs. Available values: "GRO21", "RYB21", "GRO20", "RYB20, "AMO18". For additional information, see section 2.1.1.

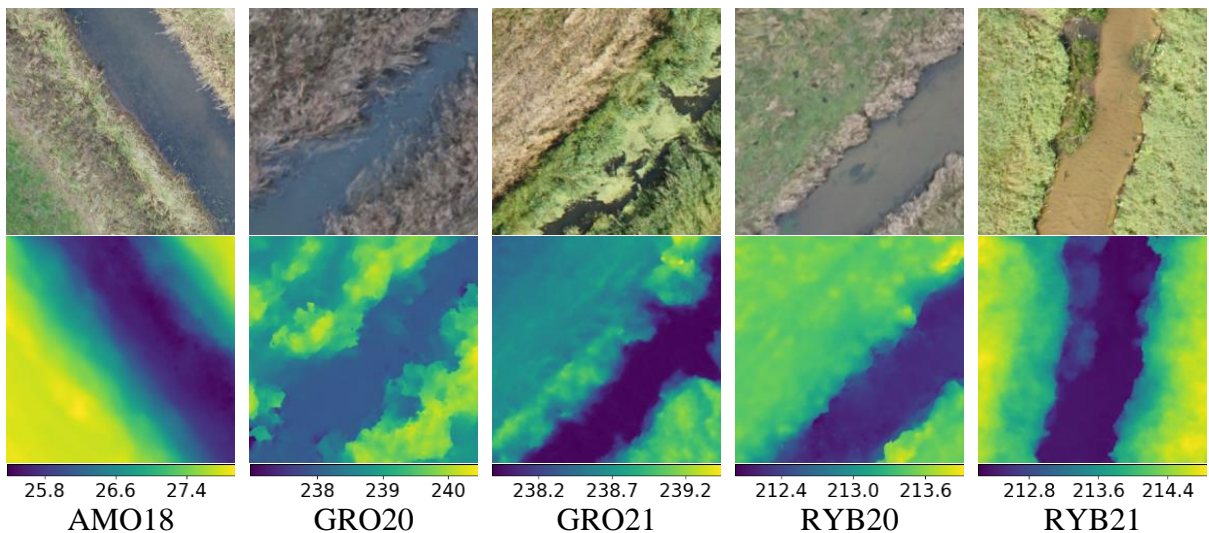

Figure 4. Examples of machine learning samples from each survey subset. Orthophotos are shown in the upper row, and DSMs (m MSL) are shown in the lower row.

## 2.2. Deep learning regression

In this work, the deep learning (DL) model based on convolutional neural networks (CNN) is used as a WSE estimator. The WSE values that are the output of the DL model are then subjected to ordinary least squares linear regression against chainage, which further improves the accuracy of the WSE estimation. Figure 5 shows the flow diagram of the WSE prediction solution.

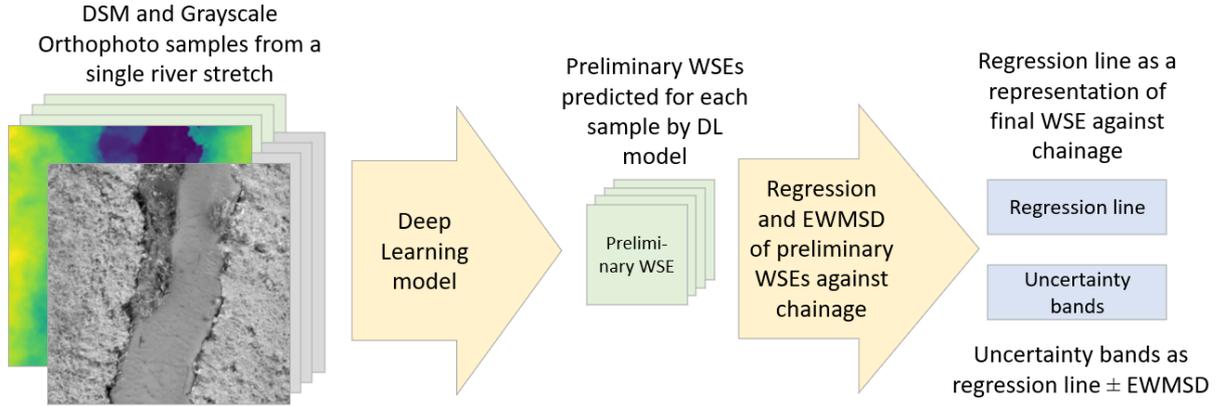

Figure 5. WSE prediction flowgraph.

*2.2.1. Tools and libraries*

The entire machine learning solution was created using the PyTorch library for the Python programming language (Paszke et al., 2019). Model training was performed using Microsoft Planetary Computer (https://planetarycomputer.microsoft.com) and PLGrid (https://www.plgrid.pl/en) computing resources. The collection of experimental results was performed using Neptune software (neptune.ai, 2020).

*2.2.2. Data preprocessing*

The machine learning solution for both training and prediction used only samples for which the difference between the maximum and minimum values of the DSM array is less than 4.5 meters. As a result, few samples containing tall trees were discarded. Since samples containing trees are sparse, they would pose a problem for the machine learning algorithm because they are outside the distribution of most samples in the data set that do not contain trees.

As the DSM and ORT arrays have values from different ranges and distributions, they are subjected to feature scaling before they are fed into the DL model in order to ensure proper convergence of the gradient iterative algorithm during training (Wan, 2019). The DSMs were standardized according to the equation:

$$DSM' = \frac{DSM - \overline{DSM}}{2\sigma}, \qquad (2)$$

where:
$DSM'$ – standardized sample DSM 2D array with values centered around 0,
$DSM$ – raw sample DSM 2D array with values expressed in m MSL,
$\overline{DSM}$ – mean DSM value of a sample ,
$\sigma = 1.197$ [m] – standard deviation of DSM arrays pixel values for the entire data set.

This method of standardization has two advantages. Firstly, by subtracting the average value of a sample, standardized DSMs are always centered around zero, so the algorithm is insensitive to absolute altitude differences between rivers. Actual water level information is recovered during inverse standardization. Secondly, dividing all samples by the same sigma value of the entire data set ensures that all standardized samples are scaled equally. It was experimentally found during preliminary model tests that multiplying the denominator by 2 results in better model accuracy, compared to standardization that does not include this factor.

Orthophotos were standardized using ImageNet (Deng et al., 2009) data set mean and standard deviation according to the equation:

$$ORT' = \frac{ORT - \mu}{\sigma}, \qquad (3)$$

where:
$ORT'$ – standardized 1-channel orthophoto gray-scale image (2D array) with values centered around 0,
$ORT$ – 1-channel orthophoto gray-scale image (2D array) represented with values from the range [0,1],
$\mu = 0.449$ – mean value of ImageNet data set red, green and blue channel values means (0.485, 0.456, 0.406),,
$\sigma = 0.226$ – mean value of ImageNet data set red, green and blue channel values standard deviations (0.229, 0.224, 0.225).

In order to increase the size of the training data set and therefore improve prediction generalization, each sample array used to train the model was subjected to the following augmentation operations: i) rotation of 0°, 90°, 180° or 270°, ii) no inversion, inversion in the x-axis, inversion in the y-axis, or inversion both in the x-axis and the y-axis. This gives a total of 16 permutations, which makes the training data set 16 times larger.

*2.2.3. CNN network*

The model used to create the supervised learning algorithm to determine a single WSE value is based on the VGG-16 architecture (Simonyan and Zisserman, 2015). The VGG-16 originally used for image classification was modified to perform single floating-point value prediction. The changes made to this model are: i) the input size of the model is 2x256x256. It is a two-channel image in which the first channel contains the DSM and the second channel is a grayscale orthophoto. ii) After a series of convolution layers, a linear transformation of the array data to a single value was applied. No activation function was used on the model output to obtain a continuous value. iii) Multi-resolution achieved by concatenation of scaled input to the output of each Max-Pooling layer.

*2.2.4. Dropout averaging*

The solution uses the dropout averaging technique (also called Monte Carlo dropout), first proposed by (Gal and Ghahramani, 2015) for uncertainty estimation. During the evaluation of the CNN model for a single sample, the prediction is performed multiple times, each time introducing random perturbations to the neural network by zeroing the random weights with a dropout layer. In this way, for a single sample, many, normally distributed, predictions are returned. The average of this distribution is the predicted WSE value for a single sample. This method improves the accuracy of the results and helps to generalize the model against overtraining. We do not extract uncertainty from this method because it is computed at a later stage (see section 2.4). In our case, 100 Monte Carlo dropout predictions were made for each sample with a weight disablement probability of 50%.

*2.2.5. Training*

The CNN network was trained with the gradient descent method using the Adam optimizer with batches of 32 samples each. A cyclic learning rate scheduler was used, which changed the learning rate in successive epochs periodically using values: $1 \cdot 10^{-6}, 5.5 \cdot 10^{-6}, 0.1 \cdot 10^{-6}$. Training was terminated when the RMSE on the validation set did not improve for 10 consecutive epochs in a row by a minimum of $10^{-3}$ m. The average training time for a single k-fold subset was about 36 minutes.

*2.2.6. Regression along chainage*

An additional increase in model accuracy was achieved by performing an ordinary least squares linear regression of the WSE results returned by the CNN model against the chainage. Regression was performed separately for each survey subset using the statsmodels Python module (Seabold and Perktold, 2010). The parameters $(a, b)$ of the fitted linear model were used for the final prediction of the WSE ($\hat{y}$) based on the formula: $\hat{y}(x) = ax + b$, where $x$ is the chainage.

Notice: please do not confuse this regression with the ground-truth data polynomial regression described in section 2.1.3.

*2.2.7. Model accuracy assessment*

The metric used to evaluate the solution was the root mean squared error. It was calculated using the formula:

$$RMSE = \sqrt{\frac{1}{N}\sum_{i=1}^{N}\bigl(y(x_i) - \hat{y}(x_i)\bigr)^2}, \qquad (4)$$

where $N$ is the number of samples in validation subset, $x_i$ is chainage of given sample $i$, $y(x_i)$ is the ground truth WSE for chainage $x_i$ (described in section 2.1.3), and $\hat{y}(x_i)$ is predicted from linear regression WSE value for chainage $x_i$ (described in section 2.2.6).

The assessment of the accuracy of the developed method was performed using k-fold cross-validation. In the data set there are 5 subsets of samples obtained in different areas and seasons. These subsets were used for k-fold cross-validation. The model was trained five times. Each time, one subset of samples collected in a given area and date was excluded from the training data set and assigned for validation. Due to the small number of samples, no testing subset was extracted.

*2.3. "Water-edge" regression*

For comparison, WSE estimates were also made using a method that did not use deep learning. For each case study, the WSE was determined using the "water-edge" method by reading the DSM values along a line routed in the water area close to the river's edge. DSM values were sampled every 0.1 m. Steep banks, trees, or grasses that grow over the water surface can contribute to erroneous WSE readings. These manifest themselves as peaks on the graph of the WSE against the chainage. These peaks were removed using the forward-backward exponential weighted moving average (FBEWMA) (Hunter, 1986). In this method, the values obtained from the forward and backward exponential moving average with a window size of 50 were averaged, resulting in a forward-backward moving average. Then the WSE values that differed from the forward-backward moving average by more than 0.1 m were removed. This process was repeated 3 times for each data subset. The optimal window size, maximum deviation value and number of iterations most likely depend on the sampling frequency, river bank slope, and amount of vegetation protruding over the river; however, the values used worked well for all case studies. A comparison of the DSM values read along the "water-edge" and centerline before and after FBEWMA filtering is shown in Figure 6. A linear regression was then fitted to the filtered values from the "water-edge" as a final prediction of the water table level.

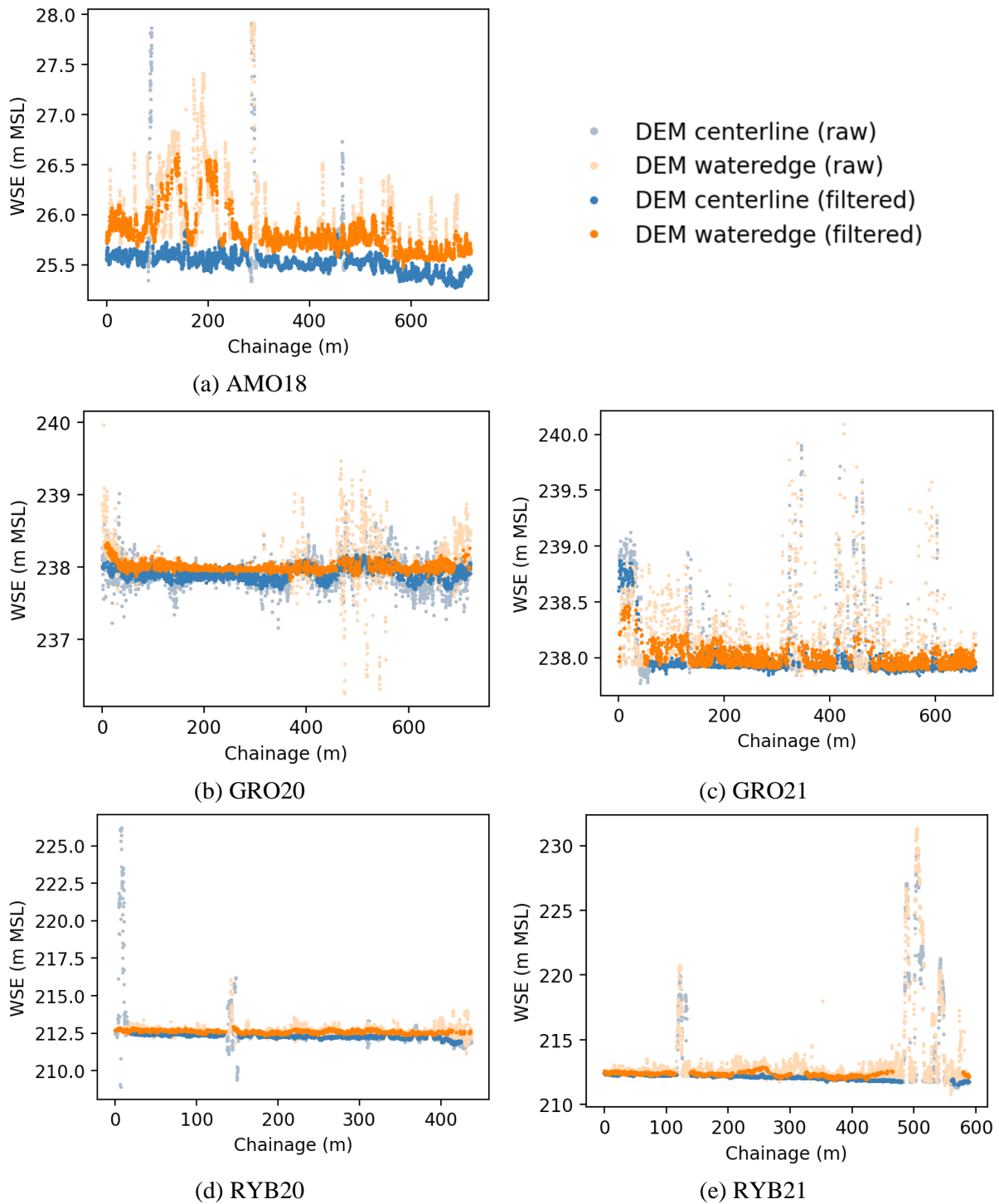

Figure 6. A comparison of DSM values read along the "water-edge" and centerline before and after FBEWMA filtering for each subset.

## 2.4. Uncertainty estimation

Forward-backward exponentially weighted moving standard deviation (FBEWMSD) (Finch, 2009; Hunter, 1986) was used to estimate the uncertainty of WSE predictions obtained by both the DL regression and the "water-edge" regression methods. An FBEWMSD window size of 10 samples was set for the points derived from the DL model, and for the "water-edge" points, the window size was 300 samples. The different window sizes are due to the different sampling densities of the two methods.

## 3. Results

*3.1. RMSE results*

Figure 7 compares the results of different WSE estimates with ground truth values. The results of the DL model predictions and the photogrammetric estimates obtained for the water-edge and centerline of the river channel are plotted against the chainage, The RMSE values for each k-fold subset of samples used for the DL predictions, are presented in Table 2.

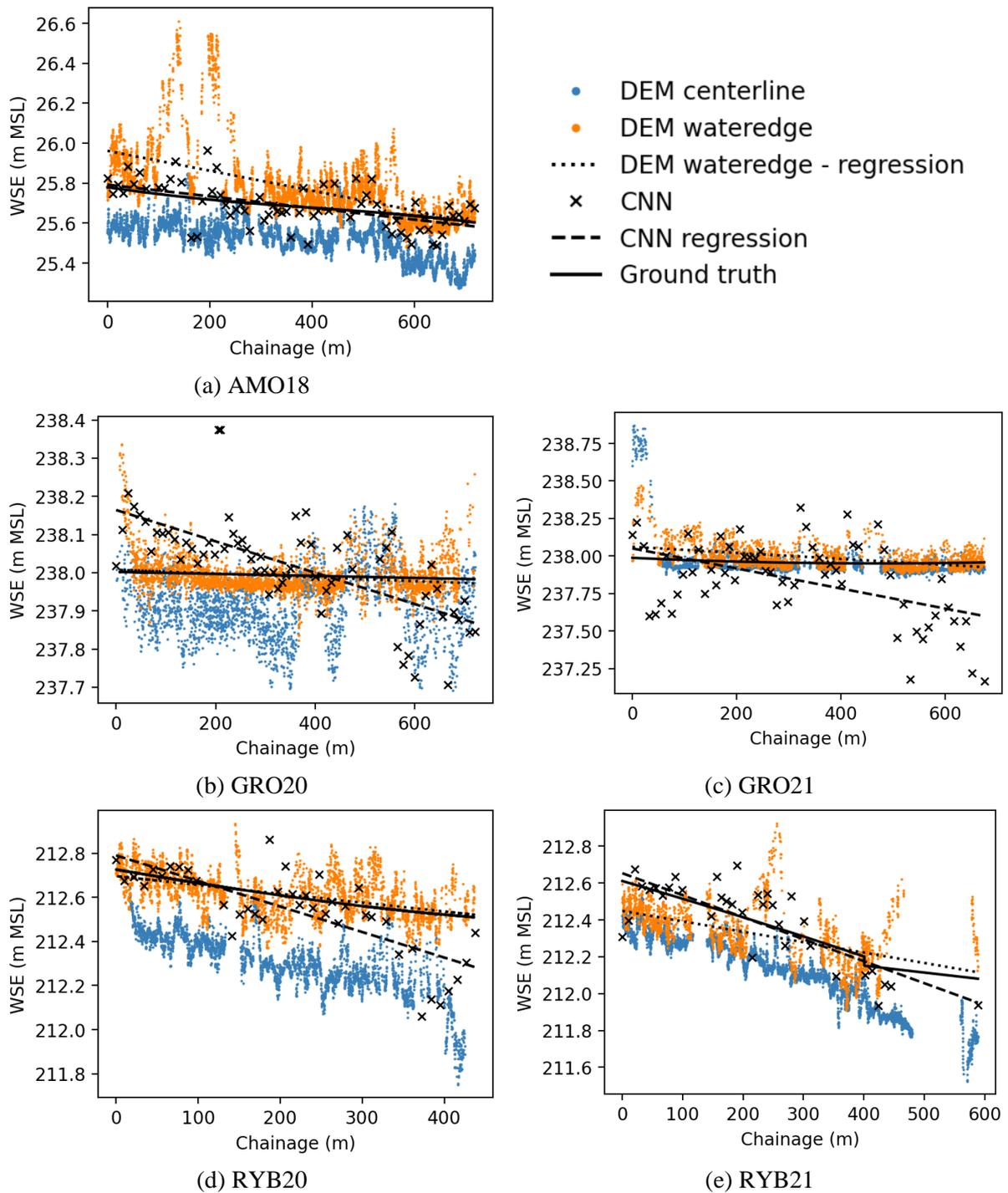

Figure 7. Ground-truth WSEs (black line), CNN WSE prediction for each sample (crosses), linear regression of CNN WSE prediction (dashed line), DSM values at the centerline filtered using FBEWMA (blue dots), DSM values at the "water-edge" filtered using FBEWMA (orange dots), linear regression of filtered DSM values at "water-edge" (dotted line).

Table 2. RMSE results achieved using different methods for each subset. RMSE results achieved using different methods for each subset. Cells are colored according to a common color scale (green – low RMSE, red – high RMSE).

| Subset | DL regression (cm) | "water-edge" regression (cm) | DL points (cm) | "water-edge" points (cm) |
|---|---|---|---|---|
| AMO18 | 1.5 | 7.7 | 9.2 | 36.7 |
| GRO20 | 8.5 | 0.4 | 12.6 | 21.8 |
| GRO21 | 17.2 | 4.2 | 29.4 | 28.8 |
| RYB20 | 10.2 | 2.2 | 15.9 | 27.3 |
| RYB21 | 3.2 | 10.1 | 12.9 | 269.7 |
| Average | 8.1 | 4.9 | 16 | 76.9 |

*3.2. Uncertainty estimation results*

The resulting uncertainty bands are visualized in Figure 8 (DL method) and Figure 9 ("water-edge" method). The average uncertainty values obtained for each subset are summarized in Table 3.

The uncertainty calibration error (UCE) (Laves et al., 2021) metric was used to evaluate uncertainty estimates. A summary of the UCE metrics obtained by the uncertainties estimated for each subset in the DL-based method and the water-edge method is shown in Table 4.

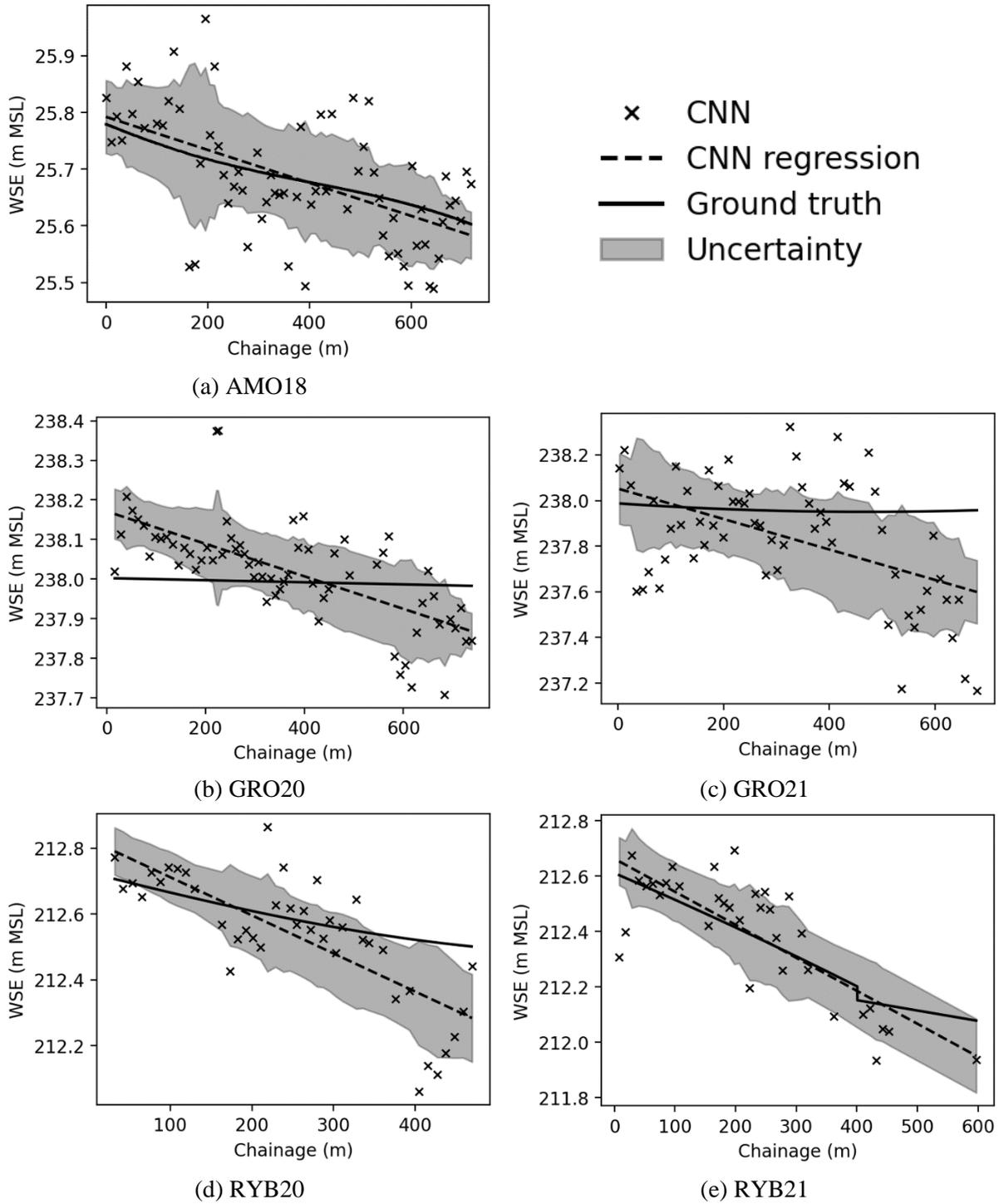

Figure 8. Uncertainty bands for the DL based method for each subset.

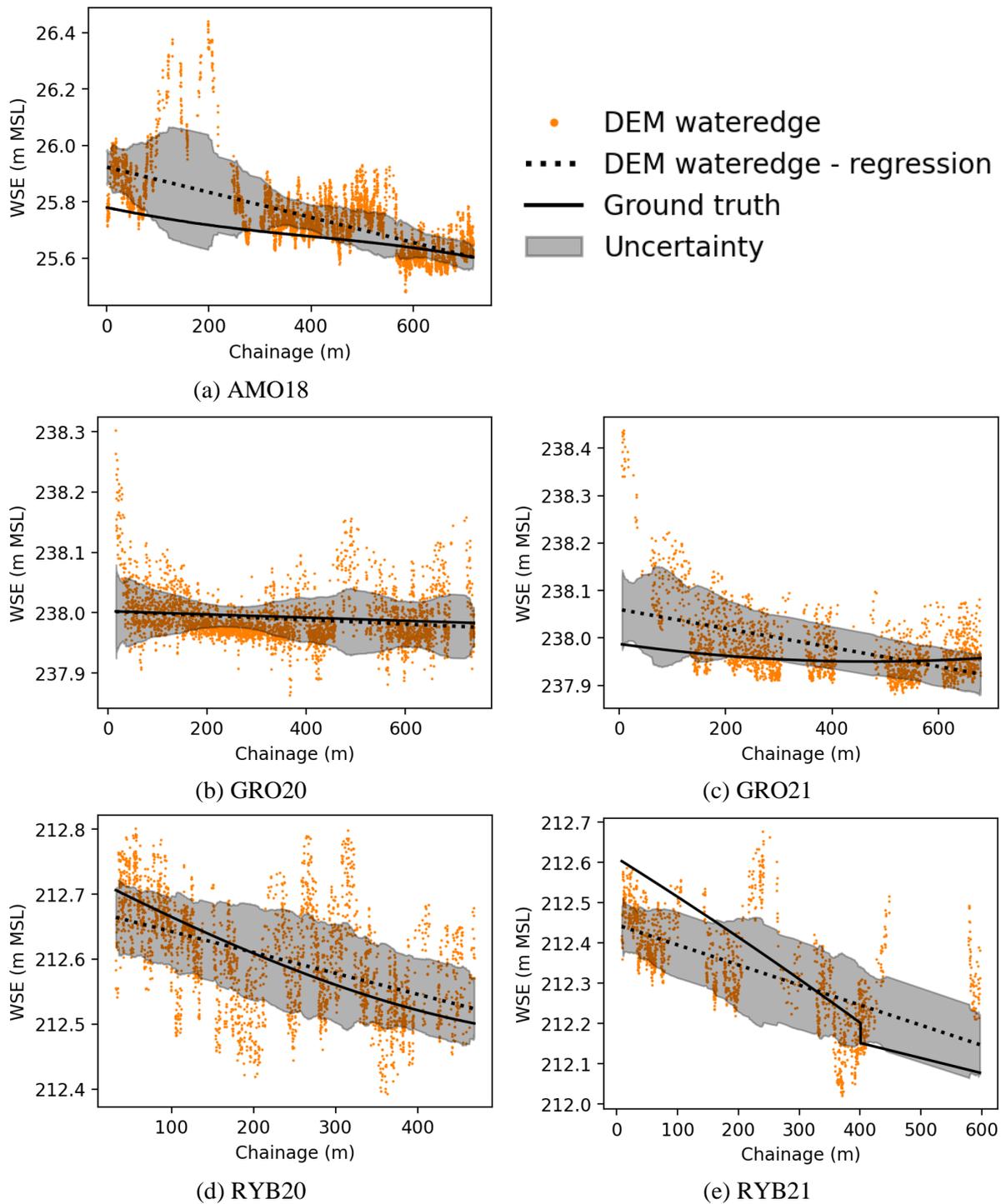

Figure 9. Uncertainty bands for the "water-edge" based method for each subset.

Table 3. Average uncertainties obtained in DL regression and "water-edge" regression.

| Subset | DL regression (cm) | "water-edge" regression (cm) |
|---|---|---|
| AMO18 | 8.4 | 7.0 |
| GRO20 | 8.6 | 3.4 |
| GRO21 | 19.6 | 5.2 |
| RYB20 | 11.5 | 6.3 |
| RYB21 | 13.0 | 8.8 |
| Average | 12.2 | 6.1 |

Table 4. UCE metric for uncertainties obtained in the DL regression and the "water-edge" regression for each subset.

| Subset | DL regression | "water-edge" regression |
|---|---|---|
| AMO18 | 9.1 | 9.2 |
| GRO20 | 8.4 | 10 |
| GRO21 | 6.1 | 9.7 |
| RYB20 | 8.3 | 9.9 |
| RYB21 | 8.7 | 8.5 |
| Average | 8.1 | 9.5 |

## 4. Discussion

Table 2 shows that the RMSE results obtained with the old "water-edge" method are unsatisfactory. This confirms the conclusion proposed by (Bandini et al., 2020) that this method is only suitable for rivers with a gentle bank slope without plants.

The results obtained from the DL regression and "water-edge" regression methods vary depending on the used k-fold cross-validation subset. On average, the RMSE obtained by the "water-edge" regression method is better than the RMSE for the DL regression. Table 2 shows that for the subsets for which DL regression performed worse, the "water-edge" regression method obtained better results, and vice versa - where DL regression obtained good results, "water-edge" regression performed poorly. This suggests that the two methods are complementary and that a solution that is a fusion of the two can be considered. A distinguishingly poor result was obtained by the DL regression method for the GRO21 subset. However, the estimated mean uncertainty for this subset is high (Table 3), which means that it has adequately served its purpose in warning of a possible misrepresentation of the predicted WSE values. The likely reason for the poor RMSE result for this subset is that it was the only one in which the water surface was covered with Lemna plants, so this feature was not considered during training. The limited ability to generalize the model's performance on data from outside the distribution is undoubtedly a shortcoming of a solution based on deep learning. To address this problem, it would be necessary to expand the training data set to include data from rivers with different characteristics.

Table 3 shows that the average uncertainties estimated in the DL regression method are higher than the uncertainties in the "water-edge" regression method. However, Table 4 shows that the uncertainty calibration error for the "water-edge" regression is higher, indicating an underestimation or overestimation of uncertainty estimates for this method. If necessary, an attempt can be made to improve the UCE metric by uncertainty calibration (Kuleshov et al., 2018; Laves et al., 2021; Levi et al., 2022).

Table 5 compares the RMSE values of the WSE measurement obtained in this article with those of other researchers. The results obtained by new DL regression and "water-edge" regression methods are predominantly outperforming existing methods based on photogrammetric products tested in this article and by (Bandini et al., 2020). Furthermore, the result obtained by the DL regression method on a Åmose Å 2018 subset proved to be better than any other UAV-based measurement method (RADAR, LIDAR, photogrammetry) tested in this case study by (Bandini et al., 2020). This is a significant success considering the small size of the training data set and the demanding k-fold cross-validation method that was used to verify the solution.

Table 5. A compilation of the RMSE values of different methods obtained in this article, by (Bandini et al., 2020) and by (Altenau et al., 2017). Arranged from lowest to highest RMSE.

| Method | Case study | Source | RMSE (cm) |
| --- | --- | --- | --- |
| "water-edge" regression | Grodzisko 2020 | This article | 0.4 |
| DL regression | Åmose Å 2018 | This article | 1.5 |
| "water-edge" regression | Rybna 2020 | This article | 2.2 |
| UAV RADAR | Åmose Å 2018 | (Bandini et al., 2020) | 3 |
| DL regression | Rybna 2021 | This article | 3.2 |
| "water-edge" regression | Grodzisko 2021 | This article | 4.2 |
| "water-edge" regression | Åmose Å 2018 | This article | 7.7 |
| DL regression | Grodzisko 2020 | This article | 8.5 |
| AIRSWOT | Tanana 2015 | (Altenau et al., 2017) | 9 |
| "water-edge" regression | Rybna 2021 | This article | 10.1 |
| DL regression | Rybna 2020 | This article | 10.2 |
| UAV photogrammetry DSM centerline | Åmose Å 2018 | (Bandini et al., 2020) | 16.4 |
| DL regression | Grodzisko 2021 | This article | 17.2 |
| UAV photogrammetry point cloud | Åmose Å 2018 | (Bandini et al., 2020) | 18 |
| UAV LIDAR point cloud | Åmose Å 2018 | (Bandini et al., 2020) | 22 |
| UAV LIDAR DSM centerline | Åmose Å 2018 | (Bandini et al., 2020) | 35.8 |
| UAV photogrammetry DSM "water-edge" | Åmose Å 2018 | (Bandini et al., 2020) | 45 |

In addition to better RMSE compared to RADAR, LIDAR and AirSWOT methods, the proposed methods also have a lower cost. The photogrammetric survey requires only a UAV with an RGB camera and a one-time GCPs delineation. Moreover, with UAVs equipped with the more accurate RTK GPS positioning system becoming more common recently, it is possible to greatly reduce the number of required GCPs (Kalacska et al., 2020).

**5. Conclusions**

The results show that machine learning techniques can be successfully used to improve existing and create new methods of WSE measurement. On average, the best result was achieved by the "water-edge" regression method. Nevertheless, where this method performed worse, the DL based method performed better, so the two methods may be complementary. Despite the small training data set, the DL solution performed well. For better generalization in more complex rivers, the training set should be expanded to include samples from more case studies.

**Code and data availability**

Geospatial raster and shape files used in the study are available online in Zenodo repository (https://doi.org/10.5281/zenodo.7185594). The source codes and the preprocessed machine learning data set can be found in the github repository (https://github.com/radekszostak/river-wse-uav-ml).

**Abbreviation list**

| | |
|---|---|
| DL | deep learning |
| DSM | digital surface model |
| FBEWMA | forward-backward exponential weighted moving average |
| FBEWMSD | forward-backward exponentially weighted moving standard deviation |
| GCP | ground control point |
| ML | machine learning |
| RMSE | root-mean-square error |
| SfM | structure from motion |
| UAV | unmanned aerial vehicle |
| UCE | uncertainty calibration error |
| WS | water surface |
| WSE | water surface elevation |

**Acknowledgements**

Research was partially supported by the National Science Centre, Poland, project WATERLINE (2020/02/Y/ST10/00065), under the CHISTERA IV programme of the EU Horizon 2020 (Grant no 857925) and the "Excellence Initiative - Research University" program at the AGH University of Science and Technology. In this study, the computing resources of the PL-Grid infrastructure were used.